\documentclass[11pt, a4paper, copyright, nonumbering]{deepmind}
\usepackage{amsmath,graphicx}
\usepackage[utf8]{inputenc}
\usepackage{color}
\usepackage[dvipsnames]{xcolor}
\usepackage{times}
\usepackage{latexsym}
\usepackage{url}
\usepackage{verbatim}
\usepackage{booktabs}
\usepackage[numbers]{natbib}
\usepackage{gensymb}
\usepackage{tikz}
\newcommand{\tikzcircle}[2][red,fill=red]{\tikz[baseline=-0.65ex]\draw[#1,very thick,radius=#2] (0,0) circle ;}%

\usepackage{algorithm}
\usepackage[noend]{algpseudocode}

\algrenewcommand\algorithmicprocedure{\textbf{function}}
\algnewcommand\algorithmicforeach{\textbf{for each}}
\algdef{S}[FOR]{ForEach}[1]{\algorithmicforeach\ #1\ \algorithmicdo}

\renewcommand{\algorithmicindent}{1.0em}

\newcommand{\algvar}[1]{\scalebox{0.85}[1]{\ensuremath{\texttt{#1}}}}

\usepackage{multicol}
\usepackage{xpatch}

\usepackage{amsmath,amssymb}
\usepackage{mathtools}
\usepackage{enumitem}
\usepackage[noabbrev,capitalize]{cleveref}
\usepackage{icomma}

\usepackage{xspace}

\usepackage{wrapfig}
\usepackage{subcaption}

\newcommand{\two}{(2)}

\newcommand{\be}{\begin{equation}}
\newcommand{\ee}{\end{equation}}
\newcommand{\bea}{\begin{eqnarray}}
\newcommand{\eea}{\end{eqnarray}}
\newcommand{\beaa}{\begin{eqnarray*}}
\newcommand{\eeaa}{\end{eqnarray*}}

\DeclareMathAlphabet{\mathpzc}{OT1}{pzc}{m}{n}

\newcommand{\datasetname}{LSVSR\xspace}

\title{Interactive decoding of words from visual speech recognition models}

\author{Brendan Shillingford \qquad Yannis Assael \qquad Misha Denil\\DeepMind}
\date{}

\makeatletter
\let\OldStatex\Statex
\renewcommand{\Statex}[1][3]{%
  \setlength\@tempdima{\algorithmicindent}%
  \OldStatex\hskip\dimexpr#1\@tempdima\relax}
\xpatchcmd{\algorithmic}{\ALG@tlm\z@}{\ALG@tlm\z@\leftmargin 0pt}{}{}
\makeatother

\begin{document}

\begin{abstract}

This work describes an interactive decoding method to improve the performance of visual speech recognition systems using user input to compensate for the inherent ambiguity of the task.
Unlike most phoneme-to-word decoding pipelines, which produce phonemes and feed these through a finite state transducer, our method instead expands words in lockstep, facilitating the insertion of \emph{interaction points} at each word position.  Interaction points enable us to solicit input during decoding, allowing users to interactively direct the decoding process.
We simulate the behavior of user input using an oracle to give an automated evaluation, and show promise for the use of this method for text input.

\end{abstract}

\maketitle

\section{Introduction}

Visual speech recognition (VSR) is the task of predicting text from the movement of a user's mouth solely relying on visual information. VSR has numerous applications ranging from silent speech interfaces to medical applications for people with speech impairments, which is a key motivating factor behind this work.
Silent speech interfaces enable speech recognition in noisy environments, or in places where one does not wish to speak audibly e.g.\ public transport and open offices.
Reliable silent speech interfaces also have enormous potential in helping  the lives of hundreds of thousands of patients worldwide. For example, in the U.S. alone, 103,925 tracheostomies were performed in
2014 \citep{hcupnet2014}, a procedure that can result in a difficulty or inability to speak (dysphonia or aphonia). Patients find their loss of voice highly frustrating and because of the difficulty of the task have to rely on professional lipreader services.
Since the vocal tract cannot be visually observed as in audio speech recognition, the mapping from phonemes to words is inherently ambiguous.  Consequently, the predictive performance as measured by word error rate (WER) is far higher.

Interactively decoding a sequence of words from a word-level model can facilitate inputting correct text despite high uncertainty. Although it is trivial to implement using beam search in word-level models,
\citet{shillingford2018large} has shown that phoneme-level connectionist temporal classification (CTC) \citep{graves2006connectionist} models are more suitable for encoding uncertainties for lipreading compared to non-phoneme CTC approaches. 
Interactive decoding in this scenario is not straightforward, as words correspond to a variable number of phonemes, and CTC marginalizes the alignment of the label sequence to the input, complicating the relationship between time in phoneme-level beam search and the number of resulting words.

In this paper, we introduce a novel search procedure for decoding words from a phoneme model, synchronized at the word level to permit the insertion of \emph{interaction points}.
The \emph{interaction points} pause the decoding procedure, at which one of the shortlisted word candidates is to be selected. This regular narrowing of the search space improves decoding quality.
Finally, we measure the performance of our method 
on a state-of-the-art system called Vision-to-Phoneme (V2P) and evaluate on a real-world dataset.

\section{Related work}

There is a large body of literature on automated lipreading which is extensively outlined in the survey papers of \citet{zhou2014review} and \citet{fernandez2018survey}.
Earlier work, including those using deep learning, has primarily focused on word classification \citep[e.g.][]{wand2016lipreading,stafylakis2018pushing,petridis2016deep}. More recent work increasingly tends to frame lipreading as a sequence prediction task, as in automated speech recognition \citep[e.g.][]{hinton2012deep,chung2016lip}.

While much investigation has focused on grapheme-based models \citep{assael2016lipnet,afouras2018deep,afouras2018lrs3,makino2019recurrent}, we follow the phoneme-based approach of \citet{shillingford2018large}, which separates the lipreading problem into a phoneme recognition module, trained via CTC loss, and a decoding module using a (weighted) finite state transducer (FST).
The phoneme recognition module consists of a series of video preprocessing steps that results in a series of frames containing the mouth region, which is input to a series of spatiotemporal convolutional layers and a recurrent neural network.
The CTC loss models the probability of a label sequence by marginalizing over alignments of the label sequence to the output of a model.
Subsequently, the decoder maps the output phoneme sequences to weighted word sequences via a lexicon and language model. Because lipreading is inherently ambiguous due to the vocal tract being visually unobservable, \citet{shillingford2018large} argue that by partitioning the problem, one can allow the recognition module to model pronunciation uncertainty and the decoder module can separately find likely word sequences consistent with the phoneme distribution. This yields an easier learning problem than having a single model modelling both sources of uncertainty, like in end-to-end grapheme-based models.

In terms of interacting with a model to make predictions, the field of interactive machine translation solves similar problems \citep{toselli2011interactive}, but usually with a focus on generating a prediction then modifying it and using active learning to learn from these corrections.
Closer to our work, \citet{harwath2014choosing} describe a method for choosing word alternates using $n$-best lists.
Such an approach is intended for displaying a single decoded sequence of words, where each word is associated with a list of alternatives to which they can be changed. In our case, however, we want the user to select a word at \emph{every} position in sequence due to the high ambiguity in lipreading.
Thus, to avoid wasted computation for unused word candidates, we only decode up to each word position, and only continue once a word is selected.

\section{Method}
\label{sec:method}

\newcommand\searchstate{search state\xspace}
\newcommand\searchstates{search states\xspace}
\newcommand\fringe{fringe\xspace}  

\begin{figure}[t]
\centering
\includegraphics[width=.5\linewidth]{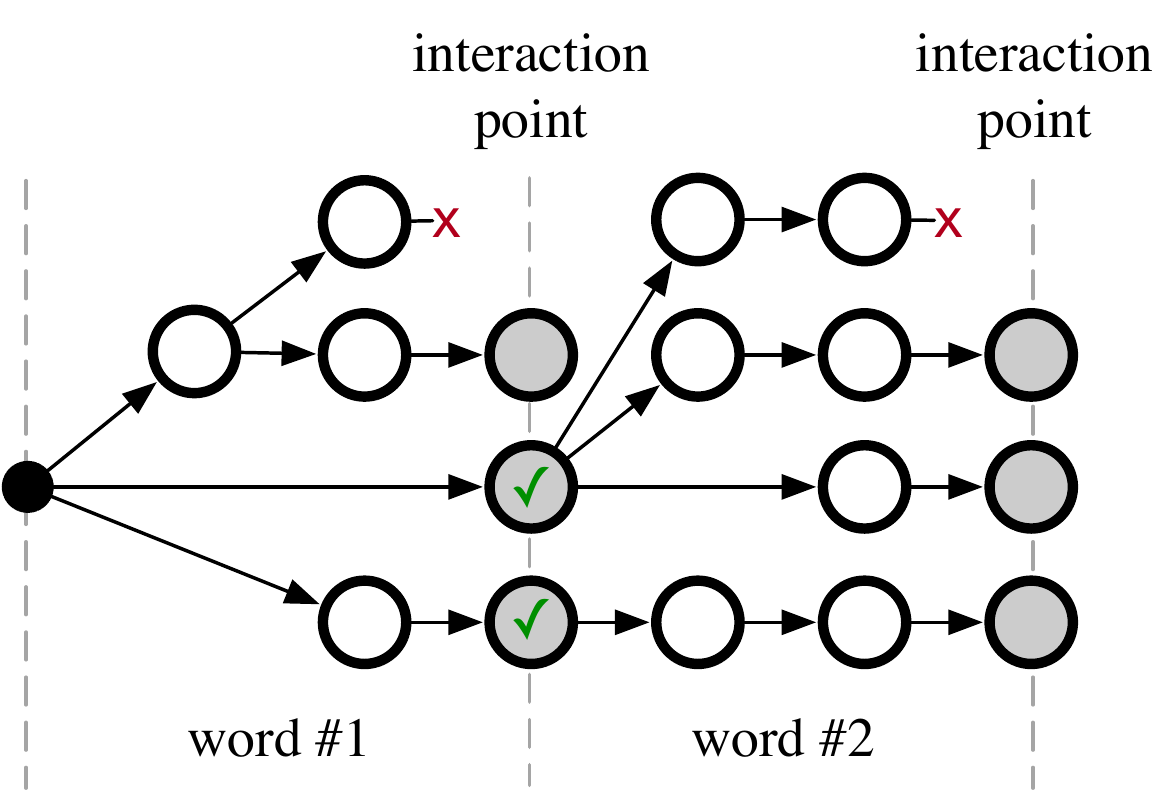}
\caption{Search is expanded at the CTC timestep level until producing enough phonemes (\tikzcircle[fill=white]{3pt}) such that the decoder FST completely outputs a word. Child search states are formed by expanding by phonemes, and for each phoneme potential branching in the FST. Some transitions lead to a dead end or get pruned due to a low score (\textcolor[rgb]{.7,0,.1}{$\!\times\!$}). Upon outputting a word, a search state is frozen (\tikzcircle[fill=gray]{3pt}) until all other search states are also frozen---due to freezing, search states may be in different CTC timesteps. Once all are frozen, an interaction point prompts for the selection of a word (\textcolor[rgb]{0,.558,0}{$\checkmark$}), and search states containing other words in this position are pruned. Then repeat to select the next word.
}
\label{fig:ph_states_word}
\vspace{-.8em}
\end{figure}

In this section, we describe how to perform a word-level search using an underlying phoneme-level CTC model. Recall that our goal is to add an \emph{interaction point} after every word, at which a single word from a list of candidates is to be selected. 
At a high level, we expand phoneme sequences recursively until every sequence of phonemes corresponds to a full word. At this point, we produce an interaction point. Then, we repeat this procedure, producing one word at a time.
See \cref{fig:ph_states_word}.

There are several challenges to realizing such an algorithm. First, there are exponentially many sequences of phonemes to which we can expand; we solve this by periodically pruning the set of \searchstates. Second, the underlying phoneme model uses CTC, so there is not a single alignment of a phoneme sequence to the model output. For this, we borrow ideas from CTC beam search that provide a rough approximation of marginalization \citep{graves2014towards}. Finally, and crucially, different words have different numbers of phonemes, and furthermore these phonemes can be output anywhere in the recognition model. 
Thus, to achieve word synchrony, unlike a standard CTC beam search, we allow different \searchstates{} to stop at different timesteps of the model's CTC probabilities.

Another important ingredient to the approach here is the (weighted) FST representing a relation of phoneme strings and word strings. FSTs are non-deterministic finite-state automata that input and output a symbol (possibly epsilon) at each arc. Each arc and accepting (final) state is associated with a weight, together determining a path's weight. We construct a \emph{decoder FST} $L\circ G$, where lexicon $L$ maps from each alternate pronunciation of a word to the word itself, and $G$ is a language model.
See \citet{mohri2002weighted} for an overview of FSTs.

The \searchstate keeps track of the phoneme sequence, the time index into the CTC per-frame label probabilities $t_{CTC}$, the FST state, the words output so far, and the cumulative FST path weight. We also keep track of quantities required for correctly computing CTC probabilities.

\begin{algorithm}[htb]
  \vspace{-0.2em}
\small
\caption{Interactive decoding for VSR. This procedure expands phoneme sequences recursively (with pruning), feeding these through the FST, until every sequence of phonemes has produced a new word. The set of words, ranked by their combined phoneme and FST (i.e.\ language model) score, is provided to the user (or the automated evaluation procedure shown in Evaluation), and the user selects one. See main text for details.}
\label{alg:interactive}
\begin{algorithmic}
\State \textbf{input:} decoder FST, \algvar{fst}; per-frame label probabilities, $\Phi$.
\Statex[0] $\triangleright$ root search state: $t_{CTC}=0$, empty phoneme and word strings, initial FST state and path weight $=0$, non-frozen.
\State $\algvar{\fringe} \gets \left[\text{root search state}\right]$
\State $\phi \gets \{ (\algvar{phonemes} = \text{`'}, t_{CTC}=0)\to(p_b=1,\,p_{nb}=0) \}$ 
\State \algvar{result} $\gets []$
\ForEach{word position $w = 1, 2, \dots$ }
  \While{non-frozen \searchstate{} exists in \algvar{\fringe}}
    \State $(\algvar{fringe}, \phi) \gets$ \textsc{ExpandFringe}(\algvar{fringe}, $\phi$, $\Phi$, \algvar{fst})
  \EndWhile
\State From $\algvar{fringe}$, build word \algvar{candidates} list sorted by score 
\State $\algvar{word}$, $\algvar{stop} \gets$ \textsc{InteractionPoint}(\algvar{candidates}) 
\State \textbf{if} \algvar{stop} \textbf{then break}
\State $\algvar{result} \gets \algvar{result} + \algvar{word}$
\State Keep \algvar{fringe} \searchstates{} with \algvar{word} at position $w$
\EndFor
\end{algorithmic}
  \vspace{-0.2em}
\end{algorithm}

The outermost loop of the search procedure (\cref{alg:interactive}) iterates over word positions. The first iteration thus finds a set of candidates for the first word. At the end of this loop, all \searchstates will be in an FST accepting state and will have output exactly one word, as illustrated in \Cref{fig:ph_states_word}.
Finally, all \searchstates that do not end in the word selected by the user are removed.
In other words, all \searchstates will now end in the word that the user selected.
Thus, at the beginning and ending of the outermost loop, all \searchstates contain identical word sequences, but potentially different phoneme sequences, $t_{CTC}$, and FST states.

\begin{algorithm}[htb]
  \vspace{-0.2em}
\small
\caption{Expand fringe by one CTC timestep. This procedure steps each non-frozen search state by one CTC timestep, some ways of which will result a new phoneme. If so, this phoneme along with each of the current FST states are input to the FST to get a new set of states.}
\label{alg:fringe}
\begin{algorithmic}
\Statex[0] $\triangleright$ Expands each non-frozen \searchstate{} by one CTC timestep,
updating CTC phoneme probabilities stored in $\phi$.
\Procedure{\textsc{ExpandFringe}}{\texttt{fringe}, $\phi$,  $\Phi$, \texttt{fst}}
\State $\algvar{\fringe}' \gets \{\}$
\ForEach{non-frozen state in \algvar{\fringe}}
  \State \algvar{phonemes} left as-is, advance $t_{CTC}$, update $\phi$ using $\Phi$
  
  \ForEach{phoneme $p$}
    \State Extend \algvar{phonemes} by $p$ advance $t_{CTC}$,
    \Statex[4] update $\phi$ using $\Phi$, add to $\algvar{\fringe}'$
    \State \algvar{new\_fst\_states} $\gets$ Feed $p$ into \algvar{fst} and
    \Statex[4] stop at accepting states
    
    \ForEach{\algvar{fst\_state} in \algvar{new\_fst\_states}}
      \State Construct new \searchstate{}, \algvar{new\_search\_state}
      \If{\algvar{fst\_state} is accepting state}
        \State freeze new \algvar{new\_search\_state}
      \EndIf
      \State Add new \algvar{new\_search\_state} to $\algvar{\fringe}'$
    \EndFor
  \EndFor
\EndFor
\State \Return $(k\text{ best \searchstates{} in }\algvar{\fringe}'), \phi$
\EndProcedure
\end{algorithmic}
  \vspace{-0.2em}
\end{algorithm}

Inside this loop, we expand each \searchstate in the \emph{\fringe} by one CTC timestep in a round-robin fashion (\cref{alg:fringe}). 
Each \searchstate is extended by every possible phoneme and advanced by one CTC timestep, and also left with its phoneme string untouched and advanced by one CTC timestep. For each of these extensions, zero or more FST states will be produced; one \searchstate is produced for each.
If the expansion of a \searchstate results in a word being output, it is \emph{frozen} and no longer participates in expansions. Once the whole \fringe is frozen, we stop expanding and insert an interaction point.

As with CTC beam search, we keep track of two quantities for a phoneme sequence at time $t_{CTC}$: $p_{b}$, the probability of the model outputting the phoneme sequence in $t_{CTC}$ timesteps and the last timestep outputting a blank symbol, and $p_{nb}$, the probability of the model outputting the phoneme sequence in $t_{CTC}$ timesteps and the last timestep outputting the final phoneme in the sequence (not blank).
Keeping these quantities separate facilitates incremental extension of phoneme sequences. 
The probability of observing a phoneme sequence up to $t_{CTC}$ is $p_b + p_{nb}$.
The update formulas computing $(p_b, p_{nb})$ at timestep $t_{CTC}+1$ from $t_{CTC}$ follow the same equations as presented in \citet{graves2014towards}, which can be thought of as a rough approximation to marginalizing out the alignment. 
As multiple \searchstates{} may share the same phoneme sequence, we store these using a hashtable keyed by (phoneme sequence, $t_{CTC}$).
For numerical stability, all operations are performed with log probabilities.

After each round-robin expansion of the \searchstates, we prune to retain a constant number of best \searchstates, as in beam search. 
The score of a \searchstate{}, used for ranking and pruning, is defined as $-\log(p_{b}+p_{nb}) + \text{FST path weight}$.

\section{Evaluation}
\label{evaluation}

Following \citet{shillingford2018large}, we train the V2P CTC model on the data from the paper as described.
We then construct a small decoder FST with a word-level bigram language model with Kneser-Ney smoothing and a vocabulary of 10,000 words \citep{ney1994structuring}. The language model was trained on the training set transcripts in \datasetname. As we only want to generate word candidate lists, and due to experimental time constraints, we only show results on a small language model. A larger language model could improve the positional ranking of the correct word.
We correspondingly subset the test set of \citet{shillingford2018large} to only the utterances that fit in this vocabulary.
We give the top 100 candidates at each interaction point, and expand up to 20 FST states per phoneme, and prune the fringe to the top 200 after expansion.

We define an \emph{oracle} that, at each interaction point, decides which word to select based on the ground-truth transcript.
The oracle is defined to mimic a human inputting text and can take 3  actions:
\begin{itemize}
\item If the word candidate list contains the current word in the transcript, the oracle selects it (\textsc{Found current}).
\item If the list contains the next word but not the current one, the oracle selects that (\textsc{Found next}).
  That is, if the current word intended to be input is not found, it is skipped and the next intended word is selected instead if present.
\item If neither the current nor the next word exist in the candidate list, the oracle picks the best scoring one (\textsc{Not found}).
\end{itemize}
The last strategy is necessary to allow the search procedure to continue even if the correct word is not present.
The oracle terminates when no words remain in the transcript.

To evaluate the method, we count the number of times each oracle action is executed. The counts and proportions of success are shown in \cref{tbl:oracle-counts}. The oracle successfully selects a word from the transcript 82.3\% of the time.
We also measure the number of times the oracle selects a word \emph{besides} the first one in the candidate list, suggesting the utility of interaction points and selecting from a candidate list instead of merely always picking the highest-probability word (\textsc{Success rate excl. first}).

\begin{table}[htb]
  \vspace{-0.5em}
  \begin{center}
    \begin{tabular}{l||r|r}
    \toprule
    \bf \textsc{Oracle action} & \bf Count & \bf \% \\
    \hline 
    \textsc{Not found} & 410 & 14.4\% \\
    \hline 
    \textsc{Found current} & 2097& 73.4\% \\
    \textsc{Found next} & 254 &	8.9\%  \\
    \hline
    \textsc{Success rate excl. first} & 1514 & 53.0\% \\
    \hline
    \bf Success rate & \bf 2351 & \bf 82.3\% \\
    \bottomrule
    \end{tabular}
  \end{center}
  \caption{Oracle actions, across 2857 interaction points. Of these, in 96 cases not counted, the last word in the utterance was not found.}
  \label{tbl:oracle-counts}
  \vspace{-0.5em}
\end{table}

We also measure the distribution of rank of the candidate words. That is, when a word is selected, at what position in the candidate list did it occur?

\begin{figure}[h]
  \centering
  \includegraphics[width=0.5\linewidth]{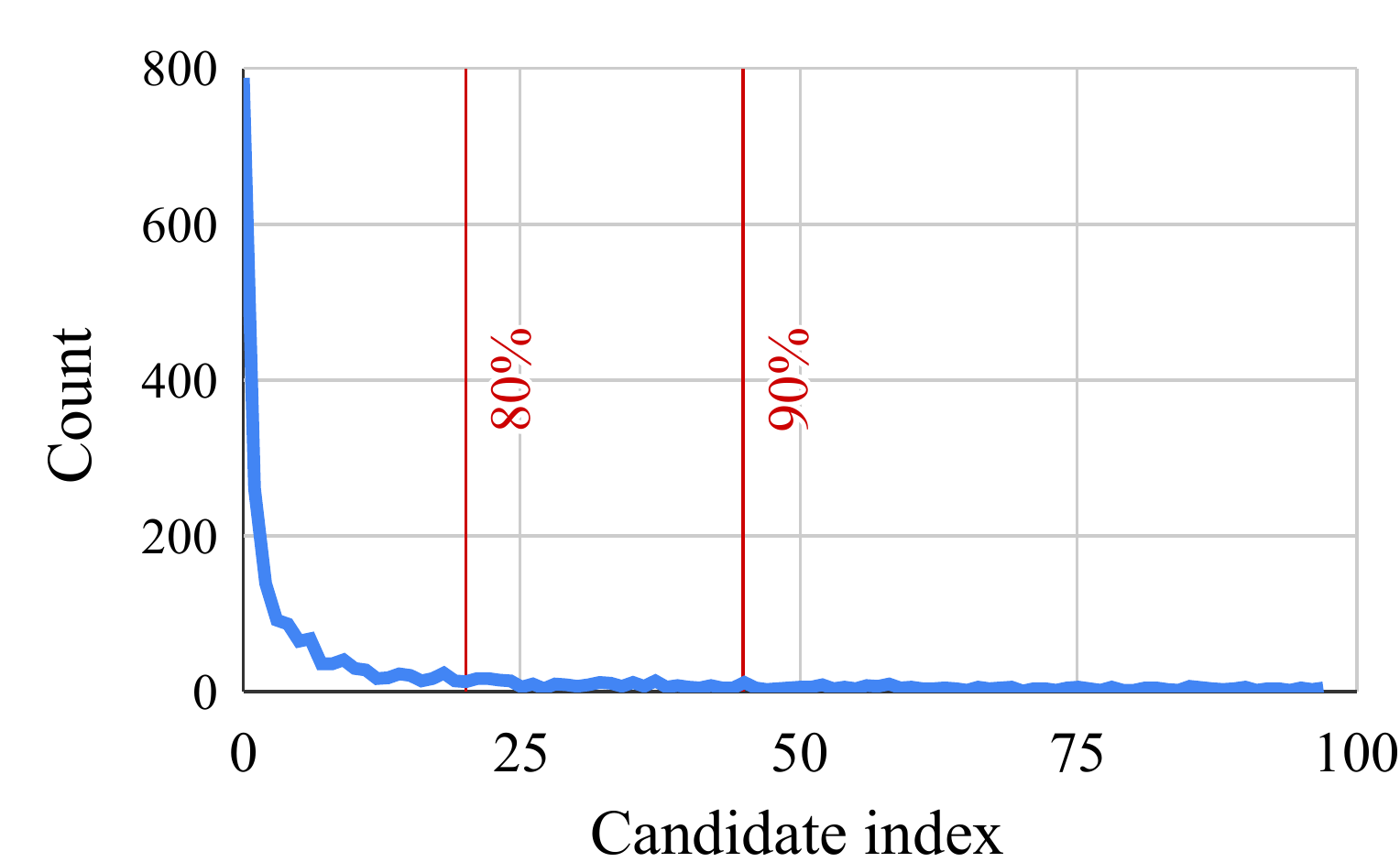}
  \caption{Distribution of selected candidate indices.}
  \label{fig:chart}
  \vspace{-0.5em}
\end{figure}

Finally, we also measure the word error rate (WER) of the predictions produced with interactive decoding using the oracle, and compare against regular decoding using the same decoder FST. For reference, we also include the performance of a large 5-gram LM (as described in \citet{shillingford2018large}) on the same data.

\begin{table}[htb]
  \vspace{-0.5em}
  \begin{center}
    \begin{tabular}{l||r}
    \toprule
    \bf Decoding method & \bf WER \\
    \hline 
    Standard (10k bigram LM) & 62.4\% \\
    Standard (5-gram LM) & 44.5\% \\
    \hline 
    Interactive (10k bigram LM) & 33.9\% \\
    \bottomrule
    \end{tabular}
  \end{center}
  \caption{WER comparison, `Interactive' denotes interactive decoding with the oracle, and `Standard' denotes CTC beam search with the FST.}
  \label{tbl:wer}
  \vspace{-0.9em}
\end{table}

\section{Conclusions}

We presented a novel method for interactively decoding word sequences from visual speech recognition models. Specifically, we described how to perform a word-level search from a phoneme-level CTC model, i.e.\ where the outer loop is over word positions, and used this to insert interaction points after each word.
We performed an automated evaluation of this procedure, which showed its promise for use in silent speech input applications.

Future work improving the quality of these results includes scaling to a larger language model, and incorporating context \emph{following} the word we are about to select. 
Furthermore, it is useful to automatically decide when an interaction point is necessary and when we are sufficiently confident that the top word is correct. Initial results showed that thresholding the score gap between the best and second-best word candidates works well.
Finally, as emphasized in this work, word-level synchronization is a useful property for interactive decoding.
This nested beam search for word-synchronized decoding from phoneme-level models can also be applied to decoding words from
grapheme-level or wordpiece-level models.

\section*{Acknowledgements}
We would like to thank Matt Hoffman and Nando de Freitas for helpful comments and feedback.

\nocite{tensorflow2016}
\nocite{numpy2006}
\nocite{allauzen2007openfst}
\bibliographystyle{IEEEtranN}
\bibliography{refs}

\end{document}